\documentclass[runningheads]{llncs}

\usepackage[T1]{fontenc}
\usepackage{graphicx}
\usepackage{booktabs}   %
\usepackage{tabularx}   %
\usepackage{multirow}   %
\usepackage{xspace}
\usepackage[
backend=biber,
style=numeric,
sorting=ynt,
maxcitenames=50,
maxnames=50
]{biblatex}
\usepackage[inline]{enumitem}
\usepackage{tablefootnote}
\usepackage{orcidlink}
\usepackage{amsmath}
\usepackage{cleveref}

\newcommand{\MIDVHOLOMETHOD}{MIDV-Holo-Baseline\xspace}

\newcommand{\ADVANCEDCLASSIFERMETHOD}{HoloVerif\xspace}
\newcommand{\WSLMETHOD}{WSL\xspace}

\newcommand{\SPANCLASSIFERMETHOD}{\ADVANCEDCLASSIFERMETHOD-Span\xspace}
\newcommand{\MSMMETHOD}{MSM\xspace}

\newcommand{\MIDVHOLO}{MIDV-Holo\xspace}
\newcommand{\MIDVDYNATTACK}{MIDV-DynAttack\xspace}

\begin{document}
\title{Temporal Modeling of Optically Variable Devices in Identity Documents}%
\author{%
Glen Pouliquen\inst{1,2}\orcidlink{0009-0002-5231-2228} \and
Joseph Chazalon\inst{2}\orcidlink{0000-0002-3757-074X} \and
Guillaume Chiron\inst{1}\orcidlink{0009-0004-3665-4900} \and \\
Oriol Ramos Terrades\inst{3}\orcidlink{0000-0002-3333-8812} \and
Thierry G{\'e}raud\inst{2}\orcidlink{0000-0002-0380-7948} \and
Ahmad Montaser Awal\inst{1}\orcidlink{0000-0002-0479-6312}%
}
\authorrunning{Pouliquen et al.}
\institute{
IDnow Research Center, Cesson-Sévigné, France \\
\email{name.surname@idnow.io}%
\and
EPITA Research Lab. (LRE), Le Kremlin-Bicêtre, France \\
\email{name.surname@epita.fr}
\and
Computer Vision Center (CVC), Barcelona, Spain \\
\email{oriolrt@cvc.uab.cat}
}
\maketitle
\begin{abstract} %
Robust remote verification of identity documents relies on analyzing faint, transparent security features like Optically Variable Devices (OVDs), or \textit{``holograms''}, within user-captured videos under uncontrolled conditions. Current systems, however, face critical limitations: existing methods often treat video frames in isolation, neglecting the intrinsic dynamic nature of OVDs and leaving systems vulnerable to swapping attacks, or focus on general holographic presence and lack the ability to verify specific OVD types.
Moreover, the economic infeasibility of frame-by-frame video annotation makes supervised training impractical.
In this work, we introduce two novel approaches for verifying the dynamic behavior of transparent OVDs protecting the holder's portrait, specifically designed for open-set scenarios where attack types are unknown during training. We demonstrate that these approaches can be trained without any attack samples in a self-supervised setting, surpassing previous state-of-the-art methods on public datasets while adhering strictly to industrial constraints.
Our results confirm that modeling temporal dynamics is essential for defeating sophisticated attacks under realistic conditions, and underscores the promise of sequence modeling and anomaly detection for OVD verification. Code is available at \url{https://github.com/EPITAResearchLab/pouliquen.26.icdar}.

\keywords{Identity Documents \and OVD Verification \and Fraud Detection \and Sequence Modeling \and Anomaly Detection}
\end{abstract}

\section{Introduction}
This paper addresses the automated verification of Optically Variable Devices (OVDs) on identity documents captured with commodity smartphones in remote Know Your Customer (KYC) scenarios.
Such scenarios are increasingly common in various applications, such as opening a bank account.
The overall process typically involves the user capturing a short video of their identity document, which is then analyzed by an automated system to verify its authenticity.
Other security measures are integrated into the process, such as liveness checks to ensure the user is present and prevents the use of static images, as well as safeguards against digital content injection, but are beyond the scope of this work.
Ultimately, this global KYC process consists in binding an online identity to a physical document, which serves as the primary source of truth for the holder's identity.

OVDs, often referred to as \textit{holograms}, are critical visible security features, particularly when they overlap and protect the photo portrait linking the document to its holder.
While current industrial pipelines rely heavily on attack detection (e.g., identifying screen recaptures or photocopies), they provide limited guarantees that the expected OVD is actually present and behaves as specified.
Furthermore, attack scenarios are inherently unpredictable, and training samples for such attacks are scarce.
his data scarcity severely restricts the performance of existing approaches in realistic settings, as demonstrated by the recent MIDV-DynAttack dataset study~\cite{pouliquen.25.icdar-midv-dynattack}, which highlights that dynamic attacks should primarily be used for evaluation rather than training.

\begin{figure}[b!]
\centering
\resizebox{0.9\textwidth}{!}{
\includegraphics[]{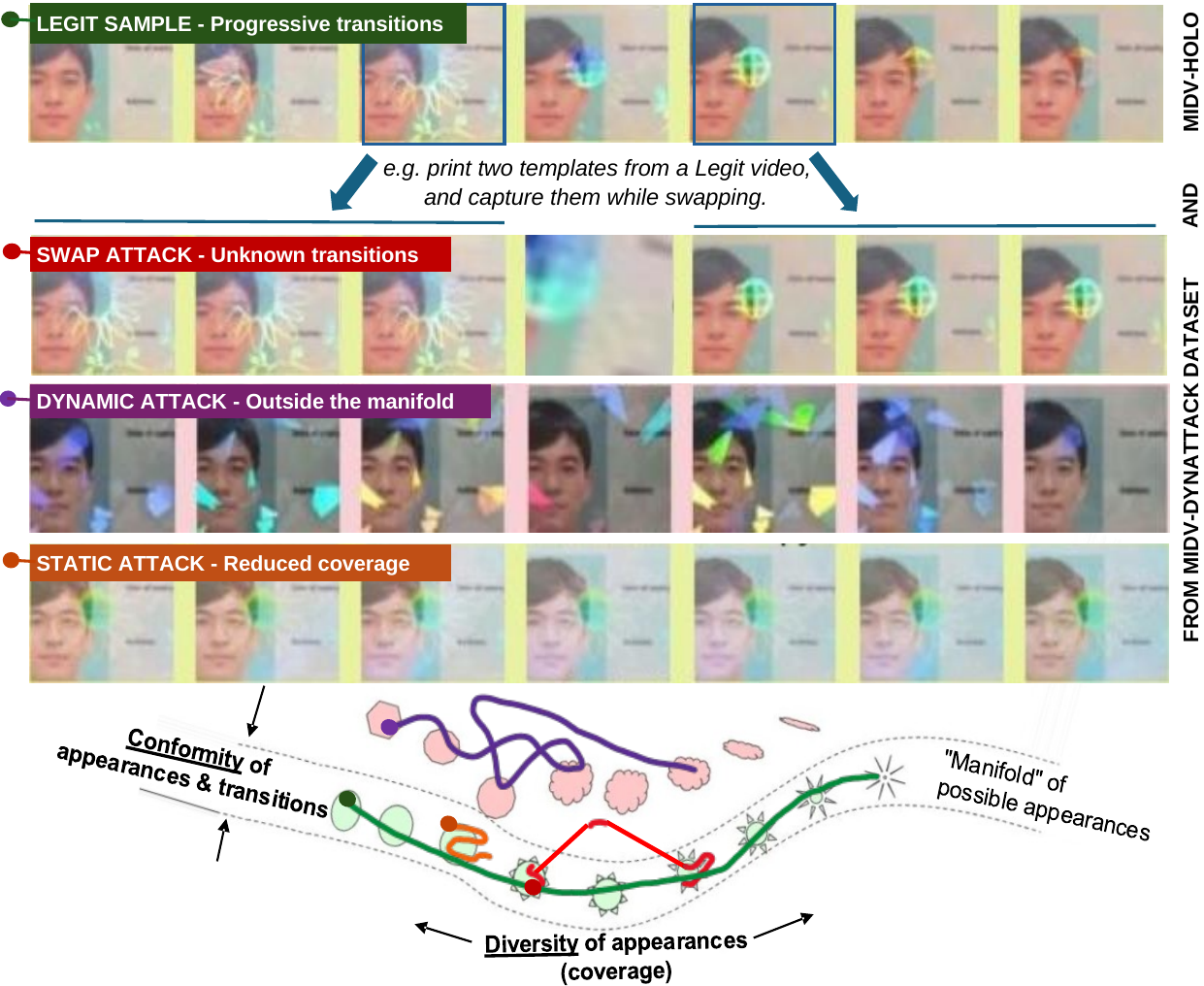}}
\caption{Importance of sequence-level modeling, and complementary directions for OVD verification to detect advanced attacks from the MIDV-DynAttack dataset~\cite{pouliquen.25.icdar-midv-dynattack}.}
\label{fig:ovdverif}
\end{figure}

In this work, we adopt a complementary perspective based on model verification rather than specific attack detection.
We aim to verify that the OVD in the portrait region is consistent with its expected visual behavior under realistic capture constraints---such as short, low-frame-rate clips from heterogeneous devices---and in the absence of fraud samples during training, which is critical for industrial applications.
As detailed in \Cref{sec:related-work}, the few existing approaches capable of handling an open attack space still rely on attack samples for training.
Moreover, they typically focus on frame-level analysis or global video features at the pixel level, failing to model the \textit{dynamics} of the OVD, which is a fundamental aspect of its security design, and thus leaving the system vulnerable to several attacks.

This work addresses these limitations with the following contributions:
\begin{enumerate*}
    \item We introduce two approaches---a discriminative span-based classifier that learns from synthetically corrupted sequences, and a generative masked sequence model that uses reconstruction error in an embedding space as an anomaly score---for verifying the dynamic behavior of transparent OVDs;
    \item We demonstrate through extensive evaluation on the MIDV-Holo~\cite{koliaskina_midv-holo_2023} and MIDV-DynAttack~\cite{pouliquen.25.icdar-midv-dynattack} datasets that these methods surpass the previous state-of-the-art while being trained in a self-supervised manner without attack samples, strictly adhering to industrial constraints.
\end{enumerate*}
Ultimately, the \emph{Masked Sequence Modeling} (MSM) we introduce aims to learn a proper manifold of some OVD's dynamics and to verify the two essential properties: \textit{conformity} to the expected visual behavior (i.e., correct color or shape changes) and \textit{variety} of appearances (i.e., sufficient change) to ensure the target is not a static counterfeit, as illustrated in \Cref{fig:ovdverif}.

The remainder of this paper is organized as follows: \Cref{sec:related-work} reviews related work in document fraud detection, \Cref{sec:approaches} details our proposed architectures and training methodologies, \Cref{sec:experiments} presents the experimental setup and results, followed by conclusions in \Cref{sec:conclusion}.

\section{Related Work}
\label{sec:related-work}

This section reviews existing approaches for Optically Variable Device (OVD) verification.
As mentioned in the introduction, we assume that the acquisition channel is secured, so that digital content injection (such as deep fakes or computer-generated graphics) cannot be directly sent for verification.
This does not prevent screens or other substitutes from being used as attack vectors, but we do not address such cases specifically as
\begin{enumerate*}
    \item 
    Presentation Attack Detection (PAD) is a mature field with established solutions for detecting such digital replays~\cite{marcel_handbook_2019}; and
    \item we believe that such approaches complement a precise verification of the expected security features of the document, in an open-set setting where exact attack implementation details are not known in advance.
\end{enumerate*}
Consequently, we focus on the subsequent challenge: detecting physical counterfeits---manipulations of the document itself and tampering with acquisition conditions---which bypass standard liveness/PAD checks.

\paragraph{Evolution of Datasets and Benchmarks.}
Public benchmarks for identity document verification have evolved from static tasks like text alteration detection~\cite{arlazarov_midv-500_2019, sibgrapi_bid} to presentation attack detection (PAD) involving printed copies or screen replays~\cite{polevoy_document_2022, park_kid34k_2023}.
However, verification of holographic security features remains under-explored compared to face morphing or splicing.
To date, the only public datasets explicitly targeting OVD verification are \emph{MIDV-Holo}~\cite{koliaskina_midv-holo_2023} and its extension \emph{MIDV-DynAttack}~\cite{pouliquen.25.icdar-midv-dynattack}, which include static and dynamic holographic attacks, respectively.
Consequently, OVD verification is a nascent research topic that is still consolidating.

\paragraph{Early Approaches Relying on Extensive Annotation.}
Early methods for hologram verification relied on handcrafted features, such as color histograms and gradient statistics, to detect expected patterns~\cite{hartl_mobile_2013}.
More recent deep learning approaches employ direct classifiers trained on frame-level annotations~\cite{chapel_authentication_2023}.
While effective on isolated, non-transparent OVDs found on some parts of banknotes or passports, such systems cannot be generalized to the transparent OVDs overlapping the portrait. These are more challenging to analyze and more critical for security, unless extensive frame-level annotations are available for training.
As a result, such supervised approaches were largely superseded in subsequent works.

\paragraph{Detection of Holographic Content Against Static Attacks.}
Shifting from modeling the collection of expected frame appearances and processing frames in isolation, two approaches proposed recovering the mask of the OVD layer of a document from a video.
Kada \emph{et al.}~\cite{kada_hologram_2022} first proposed a technique computing per-pixel color statistics over accurately registered video frames, producing a partial OVD mask for each video and enabling control over the resulting shape.
Koliaskina \emph{et al.}~\cite{koliaskina_midv-holo_2023} applied a similar technique to design a hologram detector, aiming to defeat static document copies featured in the MIDV-Holo dataset they released.
Pouliquen \emph{et al.}~\cite{pouliquen_weakly_2024} proposed a weakly-supervised contrastive framework that learns a dispersive representation of genuine OVD representations.
While providing a first, practical answer to such attacks, these approaches do not model the specific visual appearance of the target OVD nor its temporal transitions, making them susceptible to dynamic attacks such as a simple holographic layer applied to a printed copy.

\paragraph{Facing Dynamic Attacks.}
Dynamic attacks, as defined by Pouliquen \emph{et al.}~\cite{pouliquen.25.icdar-midv-dynattack}, encompass a wide range of manipulations that can be applied to the document or the acquisition conditions, such that the light reflection of the document may exhibit some degree of resemblance to the expected OVD behavior, without actually being the target OVD.
Because such attacks defeat holographic detectors, these authors proposed assessing both the \emph{conformity} of OVD appearances to the expected visual behavior and the \emph{variety} of appearances to ensure that the target is not a static counterfeit. They proposed \emph{HoloVerif}~\cite{pouliquen.25.icdar-midv-dynattack}, a method based on a pipeline comprised of background subtraction---which improves the focus on the OVD signal---, pseudo-labeling of frames likely to contain \emph{Valid} OVD content vs \emph{Non-Valid} frames which do not, and finally estimating the average of \emph{Valid} frames as a proxy for classifying the video as genuine or attack.
Despite its ability to counter a wider range of attacks, \emph{HoloVerif} treats video inputs primarily as bags of frames, neglecting the fine-grained temporal dynamics that constitute the essence of OVD security.
As a result, it remains vulnerable to swapping attacks, where several static counterfeits (with different OVD appearances) are printed and swapped during video capture, and achieves limited performance on other dynamic attacks.
Furthermore, it still relies on attack samples for training and calibration, which limits its applicability in real-world scenarios.
Our work addresses this gap by modeling the \emph{transition} between OVD appearances over time, trained exclusively on genuine samples.

\section{Proposed Methods}
\label{sec:approaches}

This section introduces two methods for video-level OVD verification designed around three shared principles: \textbf{transition awareness}, and \textbf{attack-free}, \textbf{self-supervised} training.
Unlike prior frame-level approaches, both methods process temporal frame spans (groups of 5-10 consecutive frames) to model the dynamics of holographic transitions.
Crucially, they are trained in a self-supervised manner exclusively on legitimate videos, avoiding the reliance on curated attack samples which are scarce and unpredictable in real-world scenarios.

We explore two distinct paradigms to achieve this:
\begin{enumerate*}
    \item \textbf{HoloVerif-Span}, a discriminative baseline extending HoloVerif~\cite{pouliquen.25.icdar-midv-dynattack} to temporal spans (5 frames) using a VideoMAE~\cite{tong2022videomae} backbone.
    It learns to distinguish genuine transitions from anomalies by training against synthetically corrupted versions of legitimate sequences.
    \item \textbf{Masked Sequence Modeling (MSM)}, a generative approach inspired by BERT~\cite{devlin2019bert} which leverages WSL~\cite{pouliquen_weakly_2024} as a frame encoder.
    It frames verification as an Out-of-Distribution (OOD) detection task, learning to reconstruct masked portions of valid OVD sequences and flagging reconstruction errors as anomalies, thus eliminating the need for (synthetic) negative samples during training.
\end{enumerate*}

\subsection{HoloVerif-Span --- Discriminative approach}
\label{subsec:holoverif_span}

\begin{figure}[tb]
\resizebox{\textwidth}{!}{
\includegraphics[]{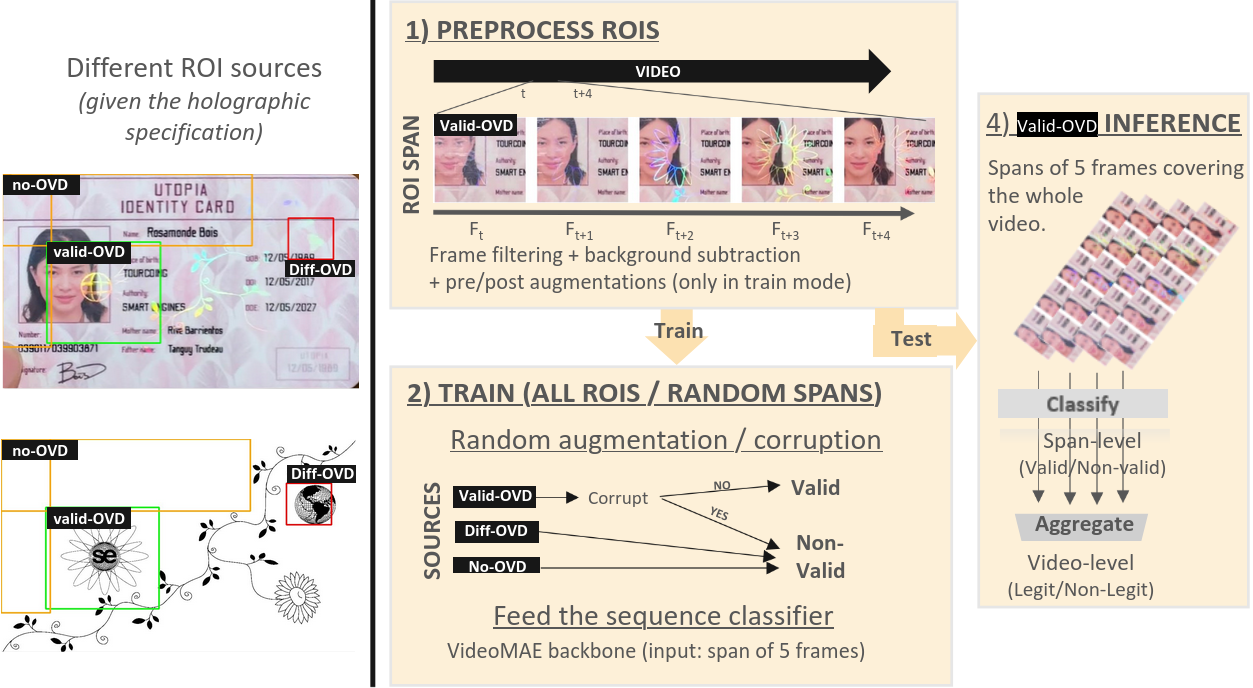}}
\caption{HoloVerif-Span overview. The OVD ROI is extracted and preprocessed via background subtraction. During training, temporal spans are pseudo-labeled as Valid or Non-Valid using synthetic corruptions, and a VideoMAE classifier is trained on these spans. At inference, a sliding window produces per-span scores aggregated into a video-level verdict.}
\label{fig:holoverifspan}
\end{figure}

\textsc{HoloVerif-Span} (\Cref{fig:holoverifspan}) is a sequence-level extension of HoloVerif~\cite{pouliquen.25.icdar-midv-dynattack} which maintains its core principle:
a binary classifier is trained on sub-video elements (\emph{spans}) that are either left intact or corrupted, and assigned pseudo-labels (\textit{Valid}/\textit{Non-Valid}) accordingly.
At inference, the final video-level verdict (\textit{Legit}/\textit{Non-Legit}) is obtained through aggregation of span-level classifier outputs to produce a single video-level score.
We preserve the overall pipeline structure, including background subtraction and asymmetric augmentation of \textit{Valid} and \textit{Non-Valid} samples, while redesigning each component to operate on temporal spans rather than isolated frames.
Crucially, these modifications enable the model to train on legitimate samples only.

\paragraph{Backbone.} The frame-level MobileViT backbone is replaced by a VideoMAE~\cite{tong2022videomae} encoder capable of capturing OVD transition dynamics directly from short clips.

\paragraph{Span Sources, Pseudo-labeling and Pre-Augmentations.}
Spans of five consecutive ROI frames are extracted from rectified videos sampled at 15\,fps.
\textit{Valid} spans are sampled from ROI regions covering the target OVD.
\textit{Non-Valid} spans originate from two complementary sources:
\begin{itemize}[leftmargin=*, nosep]
    \item \textbf{ROI-based negatives:} regions that do not contain the target OVD --- either because no hologram is present, or because the region covers a different OVD region.
    \item \textbf{Temporal corruptions of Valid spans:} synthesized by disrupting the natural temporal structure of genuine holographic transitions via
    (i)~\textit{frame repetition} --- frames are selected and repeated to simulate static or buffered content; 
    (ii)~\textit{temporal shuffling} --- frames are randomly permuted, destroying transition order; 
    (iii)~\textit{freeze simulation} --- a single frame is repeated across the full span, mimicking a static replay.
    Each strategy preserves per-frame OVD appearance while destroying temporal coherence, forcing the classifier to learn sequence-level rather than frame-level cues.
\end{itemize}
Unlike HoloVerif, where pseudo-labels are assigned at the frame level by thresholding luminance variance, pseudo-labels here are assigned at the span level based on the ROI source and corruption type.

\paragraph{Post-Augmentations.}
Background subtraction is adapted to operate on spans by computing the temporal median across the entire span, effectively isolating dynamic holographic content while suppressing the static background.
All spans then undergo a shared set of photometric augmentations designed to emulate realistic capture conditions: color jitter (brightness, contrast, saturation, hue), mild Gaussian blur to approximate motion blur and defocus, random equalization, autocontrast, sharpness adjustment, posterization, and small erased patches to simulate sensor or compression artifacts.
Synthetic shadows and localized brightness variations are additionally applied using plasma-based operators~\cite{nicolaou2022tormentor}, producing complex, spatially-varying illumination effects.
A custom transform further simulates holographic reflections in \textit{Non-Valid} spans: brightness is modulated by plasma-like fractal fields masked by random silhouettes drawn from the \emph{Quick\,!Draw!} dataset~\cite{jongejan2017quick}, producing spatially-structured highlights that closely resemble OVD optical artifacts. 
Together, these augmentations prevent the model from relying on global brightness cues while enriching it with hologram-like visual patterns during training.
\textit{Valid} and \textit{Non-Valid} spans subsequently receive asymmetric spatial treatment. \textit{Valid} spans are resized to $224{\times}224$ to preserve fine holographic patterns. \textit{Non-Valid} spans are subjected to random resized crops, flips, small rotations, and additional photometric perturbations, increasing their diversity without altering their temporal invalidity.

\paragraph{Training.}
The VideoMAE encoder (22M parameters) is fine-tuned with binary cross-entropy to classify spans as \textit{Valid} or \textit{Non-Valid}. Training uses AdamW with a learning rate of $10^{-4}$, cosine decay with a 10\% linear warm-up, and a batch size of 64. Convergence is reached in three epochs per fold.
       
\paragraph{Inference.}
At inference time, a sliding window of width 5 frames is applied over the input video sequence, yielding a per-span legitimacy score from the classifier. These scores are then aggregated via averaging to produce a single video-level legitimacy score, which can be thresholded for a binary decision or used to compute an ROC curve.

\paragraph{Limitations.} 
While this approach effectively leverages temporal information, provides a practical solution to the absence of real attacks in the training set, and serves as a comparison baseline for previous state-of-the-art, it suffers from two limitations which motivate our second, generative approach:
\begin{enumerate*}
    \item \emph{HoloVerif-Span}'s attack coverage is inherently limited by the quality and diversity of the hardcoded augmentations, weakly calibrating the boundaries around the OVD representation manifold; and 
    \item its learned representation, optimized to discriminate between \emph{Valid} and \emph{Non-Valid} spans, does not ensure a diverse representation of genuine OVD appearances.
    This limits the use of embedding diversity as a proxy for \emph{dynamic appearance variety}, potentially restricting its ability to detect simpler, static attacks.
\end{enumerate*}

\subsection{Masked Sequence Modeling --- Generative approach}
\label{subsec:msm}
\begin{figure}[tb]
\resizebox{\textwidth}{!}{
    \includegraphics[]{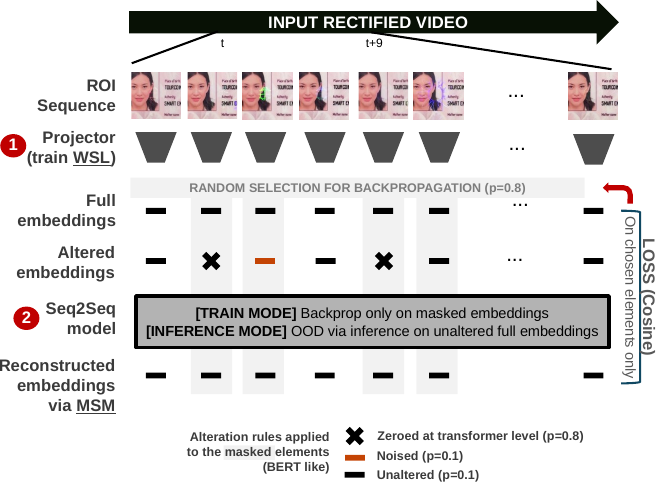}}
    \caption{Masked Sequence Modeling (MSM) pipeline. A frozen projector encodes ROI frames into OVD-focused embeddings. During training, embeddings are randomly masked and the Seq2Seq model learns to reconstruct them from legitimate sequences only. At inference, reconstruction error serves as the anomaly score.}
    \label{fig:wsl_mlm}
\end{figure}
\textsc{Masked Sequence Modeling} (MSM) is a generative approach that frames OVD verification as an Out-of-Distribution (OOD) detection problem (\Cref{fig:wsl_mlm}).
Rather than training a discriminative classifier with synthetic negatives, MSM learns to model the manifold of legitimate OVD transitions and detects attacks as deviations from this learned distribution.
This is achieved through a two-stage architecture:
a frame-level projector optimized for OVD content representation extracts embeddings,
and a transformer encoder is trained to reconstruct masked embedding spans extracted from legitimate video sequences only.
The transformer encoder provides a contextualized representation of OVD dynamics, effectively learning the underlying physical laws governing its behavior.

At inference, the method computes two complementary criteria at the span level (sequences of 10 frames), which are later aggregated at the video level:
\textbf{Reconstructability} measures how well the model (trained exclusively on spans from legitimate samples) can reconstruct a given span's representation.
It assumes that different OVD dynamics will deviate from the learned manifold and yield high reconstruction errors, hence serving as a \emph{dynamic appearance conformity} check to detect attacks with incorrect dynamics.
\textbf{Similarity} assesses the temporal diversity among frames in the learned embedding space.
It serves as a proxy for \emph{dynamic appearance variety} to detect static template attacks that might otherwise achieve low reconstruction error due to their stable nature.

\paragraph{Stage 1: Training the Projector.}
The projector serves as a frame-level encoder that captures hologram-specific visual features and produces per-frame embeddings.
To remain consistent with the ``Legit Only Training'' constraint, the projector is expected to be pretrained on a large dataset (e.g., ImageNet, CLIP) regardless of the chosen backbone (e.g., MobileViT, DINOv2), and then fine-tuned on holographic content following the Weakly Supervised Learning (WSL) approach~\cite{pouliquen_weakly_2024}.
This approach samples neighboring pairs of frames from the same video and optimizes a contrastive loss to pull their embeddings together while pushing apart embeddings from distant frames.
This encourages the model to learn a representation that captures the underlying structure of OVD appearances and their temporal evolution, while being robust to noise and irrelevant background features.
Once trained, the projector part of the pipeline is frozen.

\paragraph{Stage 2: Training the Seq2Seq Model.}
The sequence-to-sequence (Seq2Seq) model is trained using a masked reconstruction objective that forces the model to predict masked frame embeddings from the visible temporal context.
This training procedure naturally encourages the model to learn temporal dependencies between holographic states.
The key insight is that valid transitions within genuine OVDs are predictable from their context, whereas attack-induced transitions will not be, as they do not follow the temporal dynamics observed during training.

The masking strategy follows a VideoMAE-like~\cite{tong2022videomae} high masking ratio combined with a BERT-inspired~\cite{devlin2019bert} corruption protocol adapted to the continuous embedding domain.
Approximately 80\% of the embeddings in a sequence are randomly selected for alteration, while the remaining elements are left untouched and excluded from the loss computation.
Among those selected elements, 80\% are zeroed at the transformer level, 10\% are corrupted with Gaussian noise (standard deviation of 0.1), and 10\% are left unaltered.
This distribution mirrors the original BERT masking strategy adapted to the continuous embedding domain.

During training, data augmentation includes random temporal flipping ($p=0.5$), which reverses the sequence order and teaches the model that transitions are valid in both temporal directions.
The forward pass consists of applying the masking operation to produce altered embeddings and a binary mask, passing the altered sequence through the model to obtain reconstructed embeddings, and computing the reconstruction loss between the reconstructed and original embeddings on masked positions only.

The total loss combines two complementary reconstruction objectives for each frame embedding:
a Mean Squared Error (MSE) loss that penalizes absolute reconstruction differences on masked positions,
and a cosine distance loss that enforces directional similarity in the embedding space.
The total loss is computed as a weighted sum of both terms with default weights of 1.0 each.
We found that combining these two losses facilitated learning convergence.

\paragraph{Inference.}
At inference, videos are processed in spans of 10 frames, and the same projector is applied to extract frame embeddings.
The trained Seq2Seq model is then used to reconstruct embeddings for each span (without masking) and compute the two criteria:
\textbf{Reconstructability}, measured as the cosine similarity between the reconstructed and original embeddings across the span, later averaged over spans to produce a video-level score; 
and \textbf{Similarity}, computed as the mean pairwise cosine similarity between frame embeddings within the span, averaged across spans to produce a video-level score.

The final decision is made by fitting an envelope model (we use a Gaussian Model in the experiments reported here for illustrative purposes) over the joint 2D space of \emph{Reconstructability} and \emph{Similarity} on a validation set composed of legitimate samples only.
A query video is classified (\textit{Legit}/\textit{Non-Legit}) based on its likelihood of belonging to this envelope, as illustrated in \Cref{fig:oneclass_modeling}. An ROC curve can be computed by varying the decision threshold on this likelihood score.

\paragraph{Limitations.}
A key limitation of the MSM approach lies in its cascaded architecture, which stacks three distinct models: the projector (frame encoder), the sequence-to-sequence masked reconstruction model, and the final decision model.
Each stage depends on the output quality of the previous one, meaning that errors can propagate and compound through the pipeline.
In particular, if the projector fails to adequately isolate holographic content from the background or document template, the downstream embeddings will carry irrelevant or noisy information.
This degrades both the reconstruction model's ability to learn meaningful transition patterns and the decision model's capacity to discriminate legitimate from anomalous sequences.
The overall system therefore relies on the strong assumption that the projector produces embeddings that faithfully capture OVD-specific visual features which are both discriminative and robust to noise.
This is a non-trivial requirement given the weak OVD signal, the diversity of OVD designs, and the limited availability of training data for this specific domain.

\section{Experiments}
\label{sec:experiments}

\subsubsection{Methods} 
We evaluate prior state-of-the-art methods under both \textit{Legit-only} and \textit{Legit/Non-Legit} training regimes. We then focus on our two proposed methods, which are specifically designed for the \textit{Legit-only} setting---a key objective of this work, as it aligns with industrial requirements where genuine data is abundant but attack samples are scarce.

\paragraph{Existing methods (trainable from both Legits and Non-Legits):}
\begin{itemize} 
    \item \textbf{\MIDVHOLOMETHOD}~\cite{koliaskina_midv-holo_2023} computes per-pixel chromatic variance in HSV space and applies handcrafted thresholds to detect holographic activity, serving as a signal-processing baseline. No real training is performed, only calibration.
    \item \textbf{\WSLMETHOD}~\cite{pouliquen_weakly_2024} (MobileViT backbone) learns discriminative frame-level holographic representations via contrastive triplet sampling from video-level labels only.
    \item \textbf{\ADVANCEDCLASSIFERMETHOD}~\cite{pouliquen.25.icdar-midv-dynattack} (MobileViT backbone) combines temporal background subtraction with pseudo-labeling and targeted asymmetric augmentation. It currently represents the state of the art on \MIDVDYNATTACK.
\end{itemize}

\paragraph{Our proposed methods (designed for training on Legits only):}
\begin{itemize}
    \item \textbf{\SPANCLASSIFERMETHOD} (VideoMAE backbone) is a sequence-level discriminative method classifying temporal spans of five frames, with \textit{Non-Valid} spans synthesized from legitimate recordings via temporal corruptions (\Cref{subsec:holoverif_span}).
    \item \textbf{\MSMMETHOD} is a generative anomaly detection method combining the WSL frame encoder (MobileViT backbone) with a Masked Sequence Modeling reconstruction head and a single Gaussian model fitted exclusively on legitimate validation embeddings (\Cref{subsec:msm}).
\end{itemize}

\subsubsection{Datasets}
The evaluation is performed on two complementary public benchmarks:

\begin{itemize}
    \item \textbf{\MIDVHOLO}~\cite{koliaskina_midv-holo_2023}: A standard benchmark containing genuine identity documents with OVDs captured under varied conditions. It contains 700 videos --- 300 legitimate and 400 non-legitimate (mostly static template attacks). We use the identity-disjoint splits proposed in~\cite{pouliquen_weakly_2024, pouliquen.25.icdar-midv-dynattack}.
    \item \textbf{\MIDVDYNATTACK}~\cite{pouliquen.25.icdar-midv-dynattack}: An extension featuring over 1{,}200 attack videos spanning three main categories:
    \begin{itemize}
        \item \emph{Static attacks} (550 videos): Printed documents, plastic copies with or without reflections, laser and LED lighting attacks.
        \item \emph{Static-swap attacks} (200 videos): Alternation between multiple static templates to simulate holographic behavior.
        \item \emph{Dynamic attacks} (450 videos): Real holograms overlaid on fraudulent documents, including plain hologram foils, leaf patterns, double stickers, masks, and decorative holograms.
    \end{itemize}
\end{itemize}

\subsubsection{Calibration}
\label{subsubsec:calibration}
Decision thresholds are determined from the validation set which contains only \emph{Legit} samples.
We use the 95th and 99th percentiles of the score distribution on the validation set as candidate operating points, corresponding to target False Positive Rates (FPR) of 5\% and 1\%, respectively.
When \emph{non-Legit} samples are available at validation time, a ROC curve can be constructed and an operational point selected in order to balance the True Acceptance Rate (TAR) against the False Rejection Rate (FRR).
All reported AUC values use the threshold-free metric, ensuring comparability regardless of the calibration strategy.

\subsubsection{Metrics}
We report the Area Under the ROC Curve (AUC) as the primary metric, enabling threshold-independent comparison across methods and attack types. 
All experiments are repeated over 5-fold cross-validation, reporting mean $\pm$ standard deviation.
AUC values are computed per attack type and aggregated as a weighted mean over all attack videos (``Mix'' column).

\begin{table}[tb]
    \centering
    \caption{Uncalibrated results (mean AUC, 5 folds) for different training regimes (with only Legits or not) on \MIDVHOLO and \MIDVDYNATTACK datasets (test split).}
    \label{tab:resultsfull}
    \resizebox{\textwidth}{!}{%
    \begin{tabular}{lcccccccc}
        \toprule
        & \multicolumn{2}{c}{\textbf{\MIDVHOLO (L vs A)}} & \multicolumn{4}{c}{\MIDVHOLO (L) vs \textbf{MIDV-DynAttack (A)}} \\
        \cmidrule(lr){2-3} \cmidrule(lr){4-7}
        & \textbf{Vanilla} & \textbf{Photo Rep.} & \textbf{Static} & \textbf{Static-swap} & \textbf{Dynamic} & \textbf{Mix} \\
        \textbf{\#Vids. Legit/Attack $\rightarrow$} & 60L+60A & 60L+20A & 60L+110A & 60L+40A & 60L+60A & 60L+380A \\
        \cmidrule(lr){2-2} \cmidrule(lr){3-3} \cmidrule(lr){4-4} \cmidrule(lr){5-5} \cmidrule(lr){6-6} \cmidrule(lr){7-7}
        \textbf{Methods $\downarrow$} & \textbf{AUC} & \textbf{AUC} & \textbf{AUC} & \textbf{AUC} & \textbf{AUC} & \textbf{AUC} \\
        \midrule

\multicolumn{7}{c}{\textit{SOTA methods - Trained on \textbf{Legits + Non-Legits}.}} \\
\midrule
\MIDVHOLOMETHOD \cite{koliaskina_midv-holo_2023} & 93.4 $\pm$ 1.9 & 74.8 $\pm$ 2.3 & 77.2 $\pm$ 2.2 & 48.5 $\pm$ 3.4 & 8.8 $\pm$ 2.4 & 57.3 $\pm$ 1.2\\
\WSLMETHOD \cite{pouliquen_weakly_2024}& \textbf{99.2 $\pm$ 0.9} & \textbf{98.4 $\pm$ 1.1} & 95.7 $\pm$ 2.1 & 74.8 $\pm$ 8.2 & 60.7 $\pm$ 4.7 & 84.1 $\pm$ 2.9 \\
\ADVANCEDCLASSIFERMETHOD \cite{pouliquen.25.icdar-midv-dynattack} & \textbf{99.3 $\pm$ 0.6} & 94.3 $\pm$ 2.3 & \textbf{98.6 $\pm$ 0.9} & 79.2 $\pm$ 3.7 & 92.7 $\pm$ 1.9 & \textbf{94.4 $\pm$ 1.2} \\
\midrule
\multicolumn{7}{c}{\textit{SOTA methods - Trained only on \textbf{Legits}}} \\
\midrule
\WSLMETHOD \cite{pouliquen_weakly_2024} & 98.7 $\pm$ 1.0 & 97.8 $\pm$ 1.3 & 95.2 $\pm$ 2.2 & 80.6 $\pm$ 7.0 & 61.6 $\pm$ 9.1 & 84.7 $\pm$ 4.4 \\
\ADVANCEDCLASSIFERMETHOD \cite{pouliquen.25.icdar-midv-dynattack} & 52.9 $\pm$ 4.3 & 42.7 $\pm$ 4.1 & 80.9 $\pm$ 4.5 & 62.2 $\pm$ 10.9 & 73.1 $\pm$ 7.6 & 68.7 $\pm$ 5.9\\
\midrule
\multicolumn{7}{c}{\textit{Proposed methods - Trained only on \textbf{Legits}}} \\
\midrule
MSM (ours) & 94.8 ± 3.1 & 89.7 ± 6.1 & 93.5 ± 4.2 & 92.3 ± 4.5 & 85.6 ± 6.7 & 91.1 ± 3.8\\ 
\ADVANCEDCLASSIFERMETHOD Span (ours) & 80.4 ± 6.4 & 84.5 ± 4.6 & 96.5 ± 2.4 & \textbf{96.9 ± 1.7} & \textbf{98.6 ± 1.3} & 93.4 ± 2.6 \\
        \bottomrule
    \end{tabular}
    } %
\end{table}

\begin{table}[tb]
    \centering
    \caption{Calibrated results (mean Fscore, Specificity and Recall reported over 5 folds) reported on \MIDVHOLO and \MIDVDYNATTACK datasets (test split). Numbers for the SOTA methods (calibrated by Fscore maximization on the validation set) slightly differs from \cite{pouliquen.25.icdar-midv-dynattack} due to improved augmentation parameters. Our proposed methods are calibrated according to OOD thresholds of 95\% and 99\% (Legit correctly classified from the validation set). Specificities close to their respective OOD thresholds show a good generalization between splits.}
    \label{tab:resultscalibrated}
    \resizebox{\textwidth}{!}{%
    \begin{tabular}{lccccccccc}
        \toprule
        & \multicolumn{4}{c}{\textbf{\MIDVHOLO (L+A)}} & \multicolumn{3}{c}{\textbf{MIDV-DynAttack (A)}} \\
        \cmidrule(lr){2-5} \cmidrule(lr){6-8} 
        & \multicolumn{3}{c}{\textbf{Vanilla}} & \textbf{Photo Rep.} & \textbf{Static} & \textbf{Static-swap} & \textbf{Dynamic} \\
        \cmidrule(lr){2-4} \cmidrule(lr){5-5} \cmidrule(lr){6-6} \cmidrule(lr){7-7} \cmidrule(lr){8-8}
        \textbf{\#Vids. Legit/Attack  $\rightarrow$} & 120 L+A & 60 L & 60 A & 20 A & 110 A & 40 A & 90 A \\
        \cmidrule(lr){2-2} \cmidrule(lr){3-3} \cmidrule(lr){4-4} \cmidrule(lr){5-5} \cmidrule(lr){6-6} \cmidrule(lr){7-7} \cmidrule(lr){8-8}
        \textbf{Methods $\downarrow$} & \textbf{Fscore} & \textbf{Specificity} & \textbf{Recall} & \textbf{Recall} & \textbf{Recall} & \textbf{Recall} & \textbf{Recall} \\
        \midrule
\multicolumn{7}{c}{\textit{SOTA methods - Trained/Calibrated on \textbf{Legits + Non-Legits}} -- Maximizing Fscore} \\
\midrule
\MIDVHOLOMETHOD \cite{koliaskina_midv-holo_2023} & 84.8 ± 3.2 & 84.9 ± 9.6 & 85.0 ± 11.6 & 38.0 ± 20.8 & 72.4 ± 3.1 & 19.5 ± 7.6 & 3.1 ± 3.1 \\
\WSLMETHOD \cite{pouliquen_weakly_2024} & 95.8 ± 2.7 & 95.3 ± 2.8 & 96.3 ± 5.8 & 88.0 ± 12.5 & 79.3 ± 11.2 & 16.5 ± 13.5 & 4.2 ± 4.4  \\
\ADVANCEDCLASSIFERMETHOD \cite{pouliquen.25.icdar-midv-dynattack} & 95.9 ± 1.5 & 94.2 ± 3.3 & 97.3 ± 3.0 & 67.0 ± 21.7 & 92.5 ± 3.7 & 35.0 ± 7.5 & 68.9 ± 7.9 \\
\midrule
\multicolumn{8}{c}{\textit{Proposed methods — Trained/Calibrated on \textbf{Legits only} -- OOD@99 / OOD@95}} \\
    \midrule
MSM@99 (ours) & & 96.2 $\pm$ 3.4 & 72.0 $\pm$ 9.7 & 34.0 $\pm$ 25.2 & 60.7 $\pm$ 18.7 & 54.0 $\pm$ 10.1 & 58.4 $\pm$ 17.1 &\\ 
MSM@95 (ours) & & 92.6 $\pm$ 1.8 & 83.3 $\pm$ 8.8 & 56.0 $\pm$ 24.0 & 73.6 $\pm$ 13.1 & 71.0 $\pm$ 13.6 & 67.1 $\pm$ 12.0\\ 
\ADVANCEDCLASSIFERMETHOD-Span@99 (ours) & & 99.0 $\pm$ 1.5 & 19.0 $\pm$ 8.1 & 24.0 $\pm$ 2.2 & 67.3 $\pm$ 13.2 & 49.5 $\pm$ 16.5 & 87.3 $\pm$ 6.6 \\
\ADVANCEDCLASSIFERMETHOD-Span@95 (ours) & & 94.6 $\pm$ 1.8 & 35.3 $\pm$ 10.0 & 44.0 $\pm$ 10.2 & 85.1 $\pm$ 7.5 & 81.5 $\pm$ 8.0 & 94.4 $\pm$ 3.0 \\
    \bottomrule
    \end{tabular}
    } %
\end{table}

\begin{figure}[tb]
\resizebox{\textwidth}{!}{
\includegraphics[]{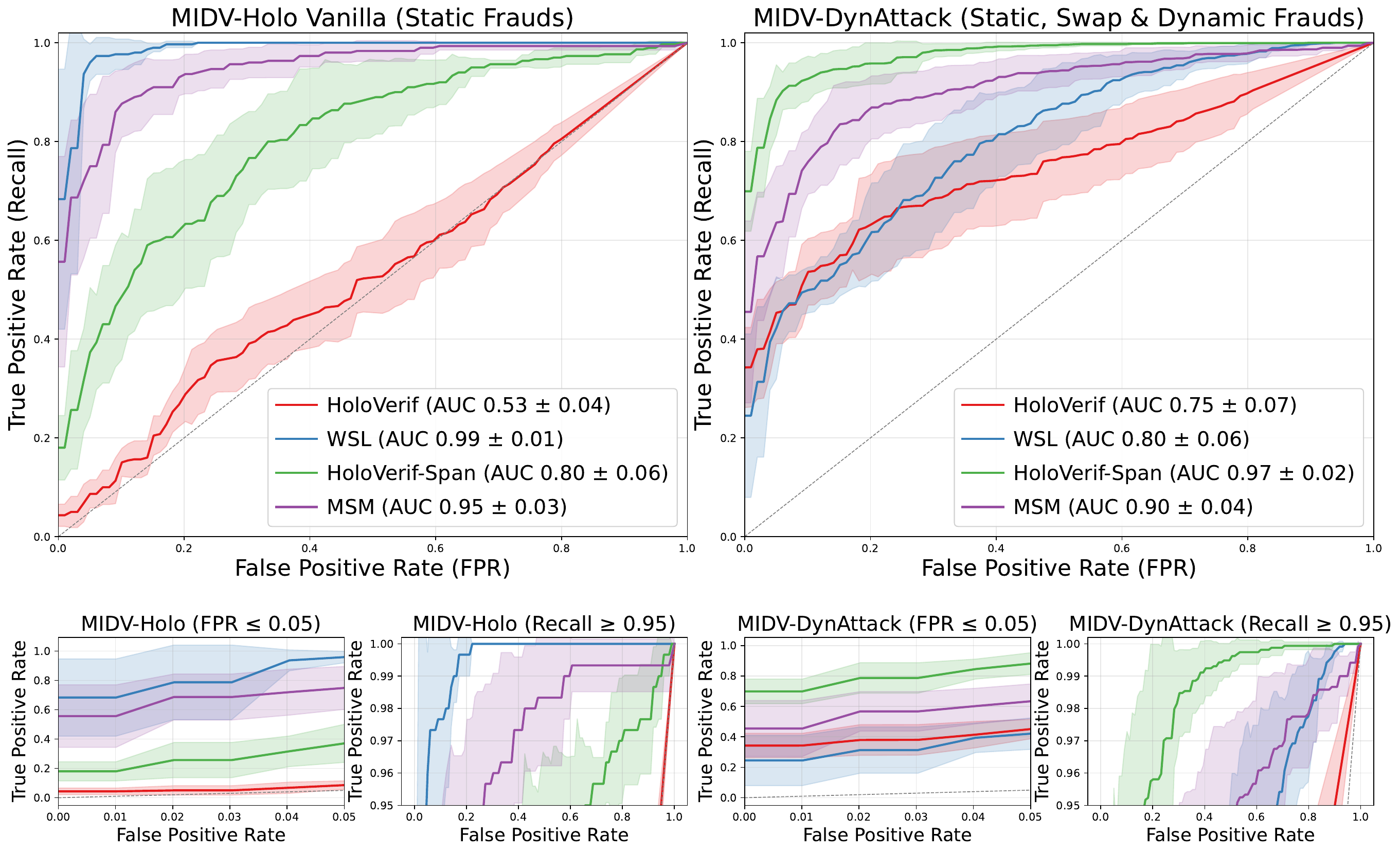}}
\caption{ROC curves for all methods trained exclusively on legitimate videos, on MIDV-Holo and MIDV-DynAttack. Top row: full ROC curves on MIDV-Holo static template frauds (left) and MIDV-DynAttack covering static template, swap, and dynamic attacks (right). Since MIDV-DynAttack contains only non-Legit samples, Legits from MIDV-Holo are used as the negative class for AUC computation. Bottom row: zoomed operating-point view at FPR~$\leq 0.05$ (left) and TPR~$\geq 0.95$ (right). Shaded bands denote $\pm$1~std over 5 folds.%
}
\label{fig:roc_curves}
\end{figure}

\begin{figure}[tb]
\resizebox{\textwidth}{!}{
\includegraphics[]{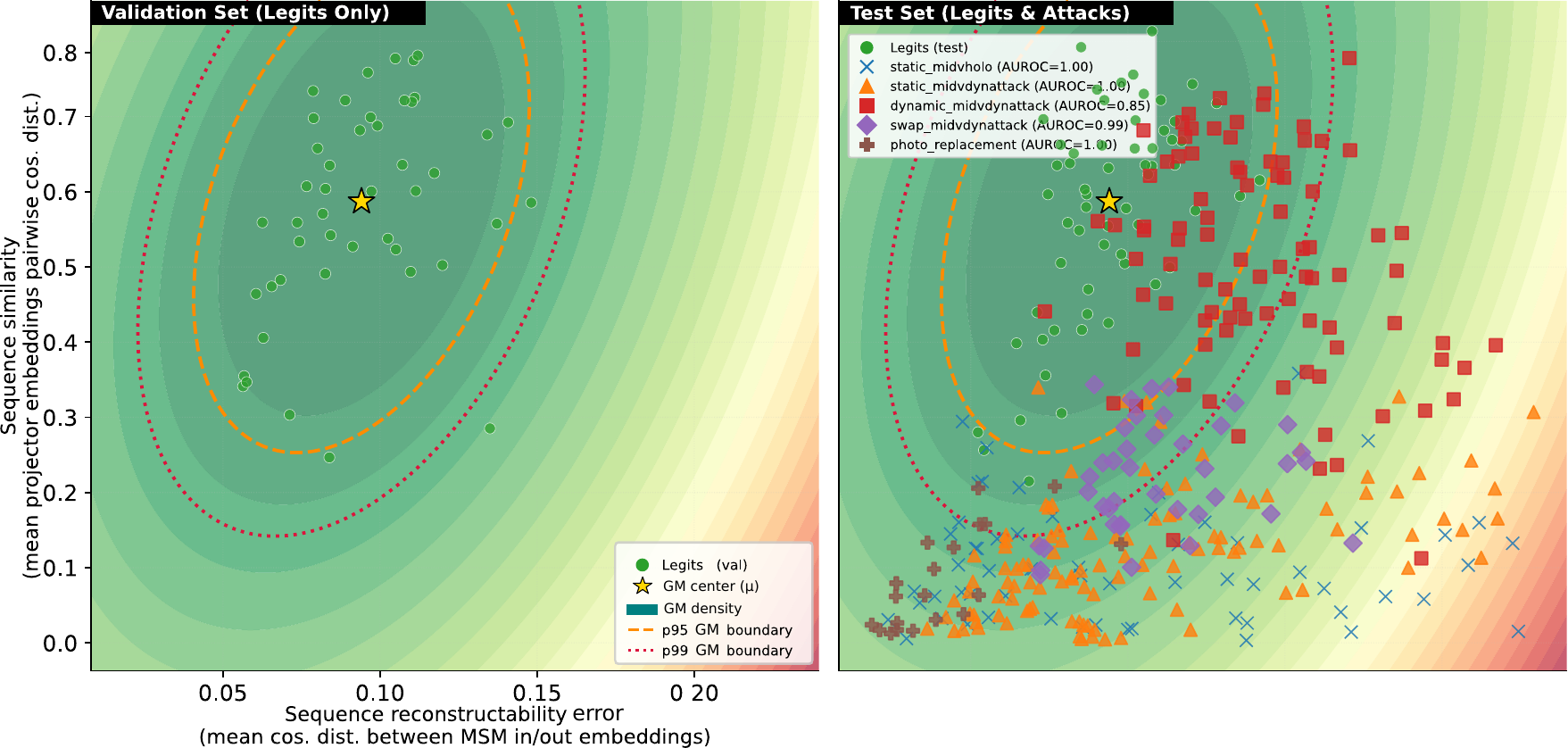}}
\caption{MSM decision space (first fold). Each point represents a video projected onto the (Reconstructability, Similarity) plane. Left: GMM fitted (k=1) on legitimate validation samples only — materialized by its density and boundaries. Right: Attack videos (from the test set) are overlaid — static attacks cluster in the low-similarity bottom region, while dynamic attacks show similar similarity but higher reconstruction error (right).}
\label{fig:oneclass_modeling}
\end{figure}

\subsubsection{Results}
\label{subsec:results}

\Cref{tab:resultsfull} presents uncalibrated results in order to compare the raw potentials of our methods versus the state-of-the-art. Since MIDV-DynAttack contains only non-Legit samples (negative class), Legits from MIDV-Holo are used as the positive class for AUC computation. \Cref{fig:roc_curves} shows the corresponding ROC curves. 
\Cref{tab:resultscalibrated} show calibrated results at given operational points in order to illustrate explicit performances (true/false rejections), aligning with the evaluation protocol defined in \cite{pouliquen.25.icdar-midv-dynattack}.

\paragraph{Performances of SOTA methods (original configuration).} When including non-Legits during training (and/or calibration), \MIDVHOLOMETHOD achieves strong performance on \MIDVHOLO Vanilla (93.4\% AUC) but collapses on dynamic attacks (8.8\% AUC): chromatic variance alone cannot distinguish genuine OVD reflections from holographic foils exhibiting similar optical signatures. \WSLMETHOD improves robustness (84.1\% overall) but similarly struggles on dynamic attacks (60.7\% AUC), where frame-level representations saturate. \ADVANCEDCLASSIFERMETHOD sets the supervised upper bound at 94.4\% overall AUC, benefiting from explicit sequence-level augmentation and background subtraction.

\paragraph{Impact of the training regime on SOTA methods.} \WSLMETHOD transfers remarkably well to the legit-only regime (84.7\% overall AUC, vs.\ 84.1\% with non-legit data), confirming that its contrastive training objective generalizes without requiring explicit fraud supervision. This makes \WSLMETHOD the most practical legit-only baseline: it requires no additional design effort and degrades gracefully. In contrast, \ADVANCEDCLASSIFERMETHOD degrades substantially (68.7\% overall AUC) when deprived of non-legit data, as its discriminative pseudo-labeling was designed to contrast legitimate and fraudulent patterns explicitly.

\paragraph{Competitiveness of our methods trained exclusively on Legits.}

Both proposed methods, trained without any fraudulent sample, achieve competitive performance against fully supervised methods. \ADVANCEDCLASSIFERMETHOD-Span reaches 93.4\% overall AUC and MSM 91.1\%, both substantially above the strongest legit-only baseline 
\WSLMETHOD (84.7\%), and closing to within one point of the supervised upper bound \ADVANCEDCLASSIFERMETHOD (94.4\%).

Table~\ref{tab:resultscalibrated} confirms these trends under real operational conditions. 
The specificities measured on the test set 
closely match the targeted OOD thresholds derived from the validation set (OOD@99: $\sim$96.2--99\%; OOD@95: $\sim$92.6--94.6\%), demonstrating 
good generalization of the calibration protocol across splits --- with a slightly larger gap for MSM. 
On dynamic attacks, \ADVANCEDCLASSIFERMETHOD-Span@95 achieves 94.4\% Recall, substantially outperforming the supervised \ADVANCEDCLASSIFERMETHOD (68.9\%), while both proposed methods significantly outperform all baselines on swap attacks.
However, recall on static template attacks and Photo Replacement remains lower, particularly at the strict OOD@99 threshold, showing limited discriminability of temporal features in this setting.

\Cref{fig:oneclass_modeling} illustrates the GM decision space of the \MSMMETHOD: static attacks cluster in the low-similarity region (near-zero intra-span variation), while dynamic attacks exhibit higher overlap with legitimate sequences, consistent with the harder detection rates in this category.

\paragraph{Capture condition sensitivity.}
The performance gap of HoloVerif-Span on \MIDVHOLO Vanilla, with an AUC of 80\% (19 point bellow HoloVerif) surfaces a broader challenge: methods trained on \MIDVHOLO legitimate videos may be sensitive to the capture conditions of that specific campaign. Future benchmarks should systematically vary acquisition devices, lighting rigs, and document nationalities to disentangle method robustness from dataset-specific biases.

\paragraph{Transition modeling as a learnable signal.}
The central finding of this evaluation is that genuine OVD temporal dynamics constitute a learnable and discriminative signal, accessible from legitimate data alone.
Notably, learning temporal dynamics finally provides a way to detect swapping attack, while strongly reinforcing the detection rate of dynamic attacks.
Frame-level methods fail on swap and dynamic attacks by design: \WSLMETHOD stagnates at 74.8\% and 60.7\% AUC even with non-legit supervision. Transition modeling closes this gap from legitimate data alone --- \ADVANCEDCLASSIFERMETHOD-Span reaches 96.9\% (+22.1 pts) and 98.6\% (+37.9 pts), with MSM independently confirming the trend (92.3\%, 85.6\%). %

\paragraph{Complementarity of Generic Holographic Detectors and Temporal Models.}
The \ADVANCEDCLASSIFERMETHOD-Span outperforms all methods on transition-based attacks --- Static-swap (96.9\% AUC, +17.7 pts vs.\ supervised SOTA) and Dynamic (98.6\% AUC, +5.9 pts) --- but underperforms on static attacks (\MIDVHOLO Vanilla: 80.4\% AUC), where no temporal discontinuity is available.
MSM, on the other hand, is consistent across all attack types (Static: 93.5\%, Swap: 92.3\%, Dynamic: 85.6\% AUC), demonstrating that a model trained exclusively on legitimate sequences can generalize to unseen fraud categories --- any deviation from the learned legitimate distribution is flagged, regardless of the attack nature.

\section{Conclusion}
\label{sec:conclusion}

We addressed Optically Variable Device (OVD) verification for remote KYC using short smartphone videos under an open-set threat model.
In this challenging scenario, fraud samples are scarce, capture conditions are uncontrolled, and the OVD signal is weak and noisy compared to the static background (suffering from transparency, spatial/temporal sparsity, and light reflections).
Previous methods either relied on heavily supervised training with attack samples, which are unavailable in production, or focused on static frame-level properties, leaving them vulnerable to swapping attacks and unable to verify the dynamic nature of OVDs.

We proposed two novel approaches for modeling OVD dynamics: \SPANCLASSIFERMETHOD, a discriminative span classifier trained with synthetic temporal corruptions, and \MSMMETHOD, a generative masked sequence model that learns the manifold of genuine OVD dynamics via reconstruction in an embedding space.
Crucially, both methods are trained exclusively on legitimate samples without any attack supervision nor manual frame-level labeling.
They demonstrate strong performance on the MIDV-Holo and MIDV-DynAttack datasets, approaching previous state-of-the-art methods that rely on attack samples for training, while finally defeating swapping attacks and exhibiting robustness across all attack types.
Overall, this confirms the importance of sequence-level temporal modeling for OVD verification and the viability of \emph{Legit-only} training.

Although the performance of proposed methods on static attacks is slightly lower than the previous \WSLMETHOD approach --- acting as a generic OVD detector --- combining both kind of approache is a straightforward way to improve robustness across the attack spectrum, as they are complementary in their detection capabilities.
Furthermore, the proposed \MSMMETHOD approach provides a unique theoretical advantage: by defining verification through \emph{reconstructability} and \emph{similarity}, it offers a unified framework that enforces both the conformity of visual patterns and the necessity of dynamic change.
This allows it to reject static counterfeits and dynamic anomalies based on principled deviations from the genuine manifold, rather than relying on discriminative boundaries learned from synthetic corruptions.

Both \SPANCLASSIFERMETHOD and \MSMMETHOD are highly promising approaches for practical deployment since they do not require anticipating specific attack implementations and can be trained on abundant production data without costly annotation.
However, the performance of both methods is still constrained by the quality of frame-level representations, which must be sufficiently focused on the OVD signal and robust to noise.
Currently, the diversity of training data is limited to a single OVD model.
Future work should focus on improving the frame-level projector and acquiring more diverse OVD data---either genuine or synthetic---to enhance representation learning, which is the main bottleneck of all approaches.
\\\\
\textit{\footnotesize This project has received funding from the European Union’s Horizon Europe research and innovation program under Grant Agreement No 101189689.}

\printbibliography

\end{document}